\documentclass[a4paper,10pt,twocolumn]{IEEEtran}%
\usepackage[numbers,sort&compress]{natbib}
\usepackage{subfigure}
\usepackage{varwidth,xcolor}
\usepackage{amsmath}
\usepackage{graphicx}
\usepackage{amssymb}
\usepackage{amsfonts}
\usepackage{algorithm}
\usepackage{times}
\usepackage{fullpage}
\usepackage{placeins}
\usepackage{epstopdf}
\usepackage{float}%
\usepackage{adjustbox}

\usepackage{afterpage}
\usepackage{booktabs} 
\usepackage{amsmath}
\usepackage{float}
\usepackage[noend]{algpseudocode}
\makeatletter
\def\algbackskip{\hskip-\ALG@thistlm}
\makeatother

\algnewcommand\RETURN{\algorithmicreturn}
\algnewcommand\PROCEDURE{\item[\algorithmicprocedure]}%
\algnewcommand\algorithmicendprocedure{\textbf{end procedure}}
\algnewcommand\ENDPROCEDURE{\item[\algorithmicendprocedure]}%
\algnewcommand{\algvar}[1]{{\text{\ttfamily\detokenize{#1}}}}
\algnewcommand{\algarg}[1]{{\text{\ttfamily\itshape\detokenize{#1}}}}
\algnewcommand{\algproc}[1]{{\text{\ttfamily\detokenize{#1}}}}
\algnewcommand{\algassign}{\leftarrow}

\algnewcommand\algorithmicswitch{\textbf{switch}}
\algnewcommand\algorithmiccase{\textbf{case}}
\algnewcommand\algorithmicassert{\texttt{assert}}
\algnewcommand\Assert[1]{\State \algorithmicassert(#1)}%
\algdef{SE}[SWITCH]{Switch}{EndSwitch}[1]{\algorithmicswitch\ #1\ \algorithmicdo}{\algorithmicend\ \algorithmicswitch}%
\algdef{SE}[CASE]{Case}{EndCase}[1]{\algorithmiccase\ #1}{\algorithmicend\ \algorithmiccase}%
\algtext*{EndSwitch}%
\algtext*{EndCase}%

\providecommand{\U}[1]{\protect\rule{.1in}{.1in}}
\graphicspath{{./figures/}}
\begin{document}
\pagestyle{plain} 

\title{SelfKin: Self Adjusted Deep Model For Kinship Verification}
\author{Eran Dahan\thanks{{Faculty of Engineering, Bar Ilan University, Israel{}}{}.
eran.dahan.ee@gmail.com.}, Yosi Keller\thanks{{Faculty of Engineering, Bar Ilan
University, Israel{}}{}. yosi.keller@gmail.com.}\\Faculty of Engineering, Bar-Ilan University, Israel.}

\date{}
\maketitle

\begin{abstract}
One of the unsolved challenges in the field of biometrics and face recognition is Kinship Verification. This problem aims to understand if two people are family-related and how (sisters, brothers, etc.) Solving this problem can give rise to varied tasks and applications. 
In the area of homeland security (HLS) it is crucial to auto-detect if the person questioned is related to a wanted suspect, In the field of biometrics, kinship-verification can help to discriminate between families by photos and in the field of predicting or fashion it can help to predict an older or younger model of people faces. Lately, and with the advanced deep learning technology, this problem has gained focus from the research community in matters of data and research.
In this article, we propose using a Deep Learning approach for solving the Kinship-Verification problem. Further, we offer a novel self-learning deep model, which learns the essential features from different faces. We show that our model wins the Recognize Families In the Wild(RFIW2018,FG2018) challenge and obtains state-of-the-art results. Moreover, we show that our proposed model can reduce the size of the network by half without loss in performance.
\end{abstract}

\section{Introduction}
The goal of Kinship Verification is to determine if two people are related and how (i.e., brothers, sisters, etc.).An automatic system that will verify the relation between two people can be beneficial in different areas. Such an automated system can help in finding the family of a known suspect; it can help determine the family of a lost child. In the field of biometrics, we can use such a concept in building a unified identity database per family and so on.\\
\begin{figure}[h]
\includegraphics[height=2.4in, width=3in]{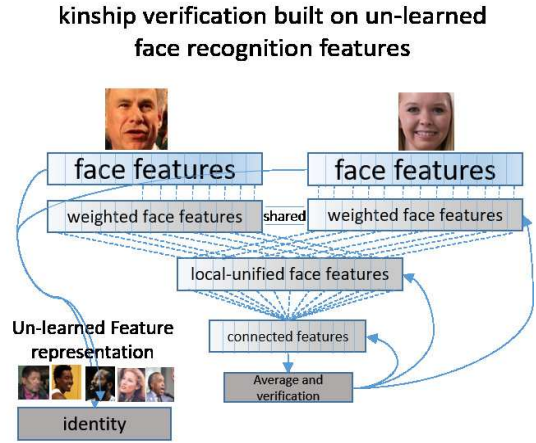}
\caption{Illustration of the method proposed in the paper, using the face features from face recognition for the task of face verification. The weighted face features are self-learned for the task of kinship verification as will be explained in the paper}
\label{fig:main_idea}
\end{figure}
Although the potential for solving this problem and Although there is an increasing interest in the computer-science community for it, the progress made so far is limited and usually implemented in specific cases and hand-picked scenarios.
Recently, the most significant dataset for kinship verification was introduced \cite{robinson2016fiw} The dataset includes different types of kinship which serves as a resource for research, the dataset was released along with a challenge (Recognize Families In the Wild, FG2018) to build a classifier for different Kinship types automatically.\\
One of the promising methods for solving the Kinship problem is with using deep learning. Deep networks can learn a different representation of faces according to a specific task (i.e., face recognition, age estimation, etc.). When investigating the Kinship Verification task one can see that we do not have a theory for known features to be extracted from the faces for this task.
That is unlike the well studied Face Recognition task, where there are known features that were considered and proven to be accurate for building Face Recognition systems.
Furthermore, we can not point what makes us decide that two people are related, so it is impossible to code this knowledge or theory to an automated system.  
The last justification was motivated us to use Deep Learning methods that can be thought by examples, and not by a theory to learn the features and classifier of Kinship-Verification problem.\\
Deep Learning methods can be implemented in different scenarios and setups. 
In the training phase, we can distinguish between a relaxed scenario and a strict scenario, in the relaxed scenario we have information about each ID (i.e., tag ID for each photo).
The ID information can be used to build our dataset, (i.e., building more negative examples or even to fine tune our face recognition model to extract more accurate face features). In literature, it is also called the image-unrestricted scenario.
In the strict scenario, the dataset is composed of only having the examples of a Kin and non-kin photos. In literature, it is also called the image-restricted scenario.\\
We can also distinguish between relaxed and strict scenarios in the testing phase.
The relaxed scenario is where we have multiple photos per ID, and we want to jointly classify them as related to some other multiple other ID's photos.
The strict scenario is where we need to make a decision only on a pair of unique photos.
In our research, we challenged the strict in the notation as explained above. The illustration of our proposed method and concept can be seen in figure \ref{fig:main_idea}\\
We can summarize our main contribution as follows:
\renewcommand{\labelenumii}{\roman{enumi}}
 \begin{enumerate}
   \item We propose a novel method to make the network self-learn the needed features to complete the task of Face-Verification. Furthermore, we show that by learning those specific features we can reduce the number of parameters that the model uses by half with almost no reduction in performance.   
   \item We propose a local and global classifier to classify the different weighted features between two feature maps; we explain why this theory is needed in the task of Kinship Verification.
   \item We train our model in the strict scenario of the Kinship-Verification problem and show that our trained model wins the RFIW2018 (FG2018) challenge.
 \end{enumerate}
The rest of the paper is arranged as follows:
Section \ref{RELATED} will summarize the previous work on the area of kinship verification; Section \ref{ARCHITECTURE}, will describe in details the architecture we propose, focusing on the feature selection layer and the local-global classifier; 
Section \ref{Dev} will include computation describing the proposed network, loss function, forward and backward calculations; 
Section \ref{ImpDet} will include training information in details and technical information for training the parameterf our network.
Section \ref{Results}, we describe our results on RFIW's, and elaborate on showing the results for different setups of the proposed network;
Section \ref{CONCLUSION} will conclude and discuss future work.

\section{Related Work}\label{RELATED}
\begin{figure}[h!]
\centering
\includegraphics[height=2.4in, width=3in]{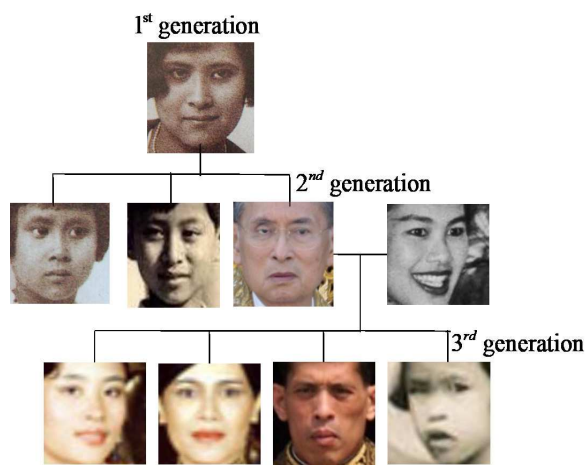}
\caption{Example of a three generation family pictures from Families In the Wild dataset}
\label{fig:three_gen}
\end{figure}

Kinship verification is a problem in the more general field of face verification and face recognition. The progress in face verification and face recognition can be divided to two main areas; the first area is the algorithm - creating more robust algorithms that describe the features of the face more distinctly. The second area is the data - getting access to a large scale of tagged data to develop more general and accurate face descriptors.\\
\textbf{Kinship datasets}: while describing the datasets that contribute to the progress of the research one can find: \cite{XIA}, there, kin-dataset was published (UB KinFace ver 2.0) the dataset consist with 600 images of parents and their children with varying ages. In\cite{Vieira2014}, a dataset of siblings - SibilingsDB was processed, the dataset includes pairs of better quality images corresponding to siblings. In \cite{FG2015Kinship}, a dataset of kinship images  (KinFaceW-I, KinFaceW-II) with different family relation types (father-son, father-daughter, mother-son, mother-daughter) was introduced and was evaluated and tested as part of the FG-2015 challenge for kinship verification. In \cite{fang2010towards} the Cornell Database Group released a dataset of 150 images pairs of parents and children (i.e., F-S, F-D, M-S, M-D). In \cite{6909623} The Family101 dataset was introduced, It contains 101 families with 607 different individuals for the total of 14,816 images. Finally, as will be described on \ref{Results} the most extensive dataset of family members kinship images made public in \cite{robinson2016fiw}, and afterward has been expanded to three-generation families images in \cite{fiwpamiSI2018} this dataset was the ground for two kinship verification challenges, RFIW2017, RFIW2018(FG2018). The diversity of the latest can be seen in figure \ref{fig:three_gen}\\

\textbf{Algorithms}: in general, it is common to divide the face verification and recognition algorithms to \textbf{three} main approaches.
\begin{itemize}
  \item \textbf{first}, engineered, hand selected features approach, where the features are engineered and tested from a database of faces and then picked to succeed in a specific task such as face verification.\\
  Using this approach we can find \cite{fang2010towards} where different features such as LBP, HOG, and Gabor, were used to encode the faces in the dataset and then selected features was used to train KNN and Kernal SVM for kinship verification. Similarly, in \cite{Dandekar} used another classifier with the LBP as face descriptor. In \cite{Mathews} computed the HMM (Hidden Markov Model) with distance features between successive edges that derived from the DCT and Sobel operator. In \cite{FG2015Kinship,Qin} used SIFT face descriptor with sparse regularized regression in the problem of kinship verification to find the essential patches from the face images, those were then classified with SVM.
  
  \item \textbf{second}, metric learning approach, where the idea is to select and learn features that can be similar for kin pairs and therefore have a higher score for similarity then non-kin pair.\\
  As in \cite{KinFACEW}, Where a descriptor was learned to cast the two images to a space where a distance vector was minimized between a kin pair. In \cite{CMML} a Cross Model Metric Learning (CMML) approach was introduced, with the use of an asymmetric scheme, where each image is processed by a different deep net that is adjusted to the input, then the distance metric is being evaluated on the outputs of the two different deep nets. In\cite{Liris-7042} TSML (Triangular Similarity Metric Learning) was proposed with multiple dimensionalities reduced descriptors such as LBP and Fisher vectors. Finally, In \cite{Haibin} DMML(Discriminative Multimetric Learning) was proposed where different distance matrices are combined by learning the optimal weights per feature,  each of those matrices is learned independently from different features.
  
  \item \textbf{third}, Deep Networks approach, deep learning seems like a promising approach for describing and extracting the features needed for classification of kin pairs, due to the deep features it can extract as shown in \cite{DEEPID,imagenet,RESNET} and due to the fact that the features describing kin pairs are not explainable (opposed to the task of face recognition).\\
  In \cite{Wang:2016} auto encoders were used to learn the representation of the difference between kin and non-kin pair, then classifying the mapping features for kin and non-kin pair. In \cite{6909623} the author introduced the idea to use deep neural encoders, to detect the most important facial features and use them as high-hierarchical features to classify between kin and non-kin pairs. Deep Learning approach was also developed by \cite{zhang12kinship}, there a softmax was used for classification and convolution layers for finding the features for representation of kin pairs.
\end{itemize}

\begin{figure}[H]
\includegraphics[height=1.6in, width=3in]{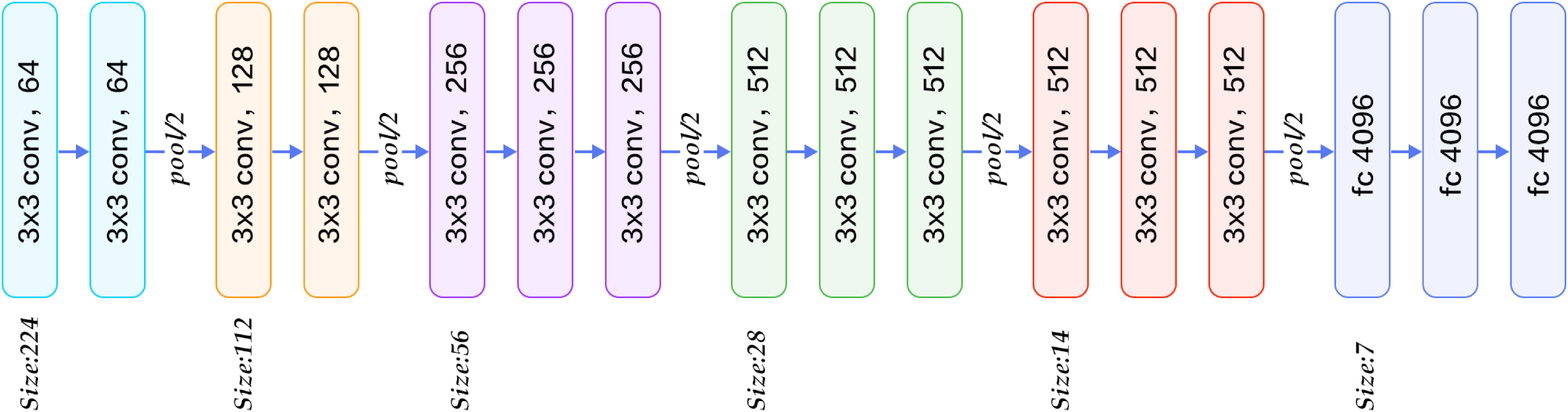}
\caption{Architecture of VGG-16}
\label{fig:vgg16}
\end{figure}
  
\section{Architecture Of SelfKin Network}\label{ARCHITECTURE}
In this section, we will start by describing our Architecture in general. Next, we will be focusing on each part of the Network and explain what is its rule in the solution of kinship verification.
\subsection{Architecture Overview}
The architecture of the SelfKin network composed of the idea to address a solution to two main challenges.
On the one hand, we would like that the representation (the features) of the images, fed to the network, will be as general as possible and not person-specific, so it will be possible to separate between kin and non-kin pairs in different poses, illuminations, etc.
On the other hand, we would require the representation to be flexible enough such in the face recognition task, so we will be able to learn an exact description of the faces in respect to some standard features of kin pairs.\\
As will be shown in the next subsections, the architecture of SelfKin network solves the need to compose features that were made for face recognition in a way that is more suitable for kinship verification. 
As shown in \ref{fig:total_arc} our Architecture is constructed with four main blocks. Face recognition features, extracted per person, Self-weighted feature selection block, Local-Global classifier and Global averaging layer.

\subsection{Face Recognition Features}
During face recognition task, we represent each face as a set of features that can then be used to classify different identities.
It has been shown that those features are also invariant to different pose, illumination, gender, etc.
Those features are constructed based on all the face examples shown to the network.\\
Since those features are generalized across different faces and shown to be suitable to describe the face for a specific person, we will use them as descriptors for faces.
We will then need to compose those features in a way that will be possible to use them to classify kin and non-kin pairs.\\
For the face extract feature task, in the architecture that is proposed here, we used VGG-FACE \cite{Parkhi15}, VGG-FACE is a deep network with the architecture of VGG, composed of convolution, maxpool and fully connected layers, the architecture of this network can be seen in figure \ref{fig:vgg16}.
This network has been trained and evaluated for the face recognition task with a dataset of 2,622 celebrities containing a total of 982,803 images from the web, the dataset \cite{GBHuang}, is an open dataset for use.\\
To agree with the rules of the RFIW2018 it should be noticed that there are two different possible setups for the challenge, those setups influence the design of this network. The restrict setup - which means that the ID of the person is not known.
The unrestricted scenario where there is a label of ID for each photo.
In this work, we aim to solve the restrict mode of the challenge, so we did not fine-tune the VGG-FACE network with identities of the dataset.\\
The output of VGG-FACE is classification to one of the 2,622 identities, and the embedding layer size is a feature vector with 4,096 features.
For the task of kinship verification, we will extract the 4,096 features for each image in the questioned pair. 

\subsection{Self-Weighted Features Layer}
The features extracted from the face images are features learned for the use of face recognition.
Those features are learned by tuning the network to recognize a specific person and ideally separate from one person to another.
This descriptor is not necessarily needed to classify kin and non-kin pairs and can be too informative and not generalize well for the kinship verification problem.\\
To overcome the un-generalization of the description for kinship verification we suggest inserting a self-adjusted weighting feature layer. This layer is learned during the task of kinship verification and aims to give high weight value to features that are needed for the task of classification and reduce the weights for the features that are not needed.\\
After training this layer we can threshold the weights and cancel features that do not contribute to the kinship verification task.
Not only it will help the network to focus on the right features it will also help to reduce the total number of weights the network has.
\begin{equation*}
\begin{split}
 &{{X}_{\text{features}}} = {{x}_{1}},{{x}_{2}}...{{x}_{n}} \\ 
 &{{X}_{\text{masked}}} = {{W}_{\text{mask}}}\cdot{{X}_{\text{features}}} = {{w}_{1}}{{x}_{1}},{{w}_{2}}{{x}_{2}}...{{w}_{n}}{{x}_{n}} 
 \end{split}
\end{equation*}

The weights of this layer $w_1\dots w_n$ are trained, end to end, with training the classifier of the network.
In order to preserve the symmetry of the problem, the weights of this network are shared between the pair of images.
When thresholding the weights, we choose a value of the threshold for all the weights in this layer and cancel the weights (e.g. assign the value of $0$) below this threshold.
The later leaves us with fewer features and fewer weights for the next layers.

\subsection{Local-Global Classifier}
\begin{figure}[H]
\centering
\includegraphics[height=2.4in, width=3in]{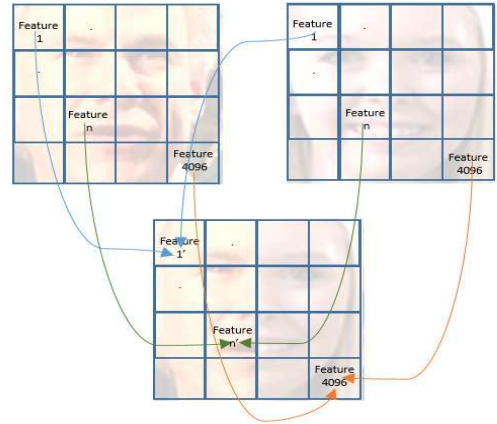}
\caption{Illustration of the local features convolution layer, each feature in the first image is weighted and summed with the corresponding weighted feature from the second picture. The resulting new feature is a nonlinear function of the above sum}
\label{fig:local_features}
\end{figure}

Based on the weighted features gained from the face descriptor, we propose to develop a local and global classifier.
The local part of the classifier will convolve the matching features (e.g. by location) between the images, making a composed feature-map from the two separate feature-maps.
Mathematically, the local part of the classifier will take a nonlinear weighted sum of each two matching features to produce a new combined features map. An illustration of this part can be seen in figure \ref{fig:local_features}.

\begin{figure*}[t!]
\includegraphics[height=2.2in, width=6.2in]{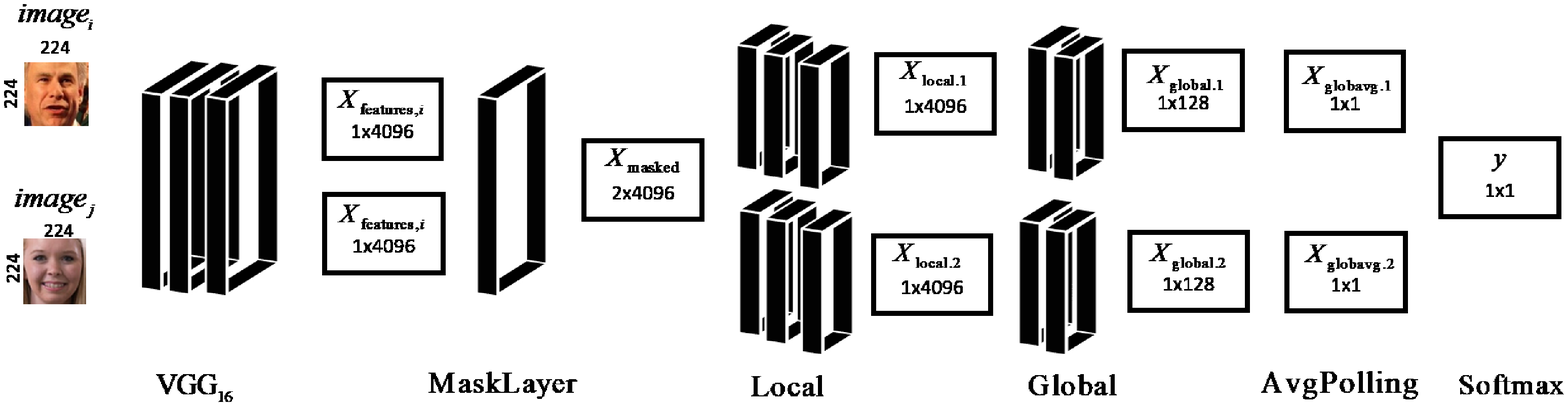}
\caption{Network Architecture and Flow, the two images are input to the VGG-FACE, the output is face features that are then self weighted and classified}
\label{fig:total_arc}
\end{figure*}

As opposed to a "regular" convolution layer where the different weights of each filter are shared through the image, this proposed layer has local weights that are specific for each pair of features by there location. In this way, we can look at each of the new features as a composed nonlinear sum of the two original features. The weights of the nonlinear combination are adjusted with respect to the kinship verification task. To preserve the symmetry of the problem we are not adding a bias term to this layer.\\
By forcing the same one weight per original feature location, we are causing the network to effect the same way to all the same location features across the images during training.\\
As described above, the motivation to do this local convolution is to create one combined feature and train it in the task of kinship verification.
The local part of the classifier preserve the original size of the feature maps (e.g. if we had $4,096$ features per image, then the size of new feature map will also be $4,096$)\\
The global part of the classifier is a fully-connected layer that computes a nonlinear combination of the new fused feature map. This layer is aimed to make a combination of spatial features in the new feature map.
This part of the classifier decreases the size of the feature map to the size of the output of the fully connected layer, in our case, it will be $128$.

\subsection{Global Averaging Layer}
After applying the Local-Global classifier, the new feature map is a  128-features vector.
Following \cite{global_avg}, we use global averaging layer.
This layer helps to prevent overfitting the training data and minimizing the size of the feature-vector to classify it.
To correctly describe our classes (Kin, Non-Kin) we use two different sets of weights in the Local-Global classifier, one for each class. As a result, the outcome of the Local-Global classifier is two feature maps, one for each class.
To make a binary decision, we average the $128$ feature vector for each class and use softmax to classify it to a kin or non-kin pair.

\section{Derivative And Loss Function}\label{Dev}
In this section we compute the forward pass values and develop the derivative of the network, we will describe the loss function of the network. 
\subsection{Forward Pass Computation}
The input to the network are two face images $224$ by $224$ pixels by $3$, we will assign them the notation $i\in (1,2)$, following the architecture is shown in figure \ref{fig:total_arc} the forward pass can be computed as:

\begin{equation*}
\begin{split}
&{{X}_{\text{features},i}} = {{x}_{1,i}}\ldots {{x}_{4096,i}}=\,VG{{G}_{16,\text{FACE}}}(\text{image}{_{i}})\\
&{{X}_{\text{masked},i}} = {{W}_{\text{mask}}}\centerdot {{X}_{\text{features},i}} = {{w}_{1}}{{x}_{1,i}}\ldots {{w}_{4096}}{{x}_{4096,i}}\\ 
&\phantom{\,\,\,\,\,\,\,\,\,\,\,\,\,\,\,\,\,\,\,\,\,\,\,\,\,\,\,\,\,\,\,\,\,\,\,\,\,\,\,\,\,\,\,\,\,\,\,\,\,\,\,\,\,\,\,\,\,\,\,\,\,\,\,\,\,\,\,\,\,\,\,\,\,\,\,\,\,\,\,\,\,\,\,\,\,\,\,\,\,\,\,\,\,\,\,\,\,\,\,\,\,\,\,\,\,\,\,\,} i\in (1,2)\\
&{{X}_{\text{masked}}} = 
 \begin{bmatrix}
   {{X}_{\text{masked},1}} \\ 
   {{X}_{\text{masked},2}} \\ 
 \end{bmatrix}\\
&X_{\text{local},j} = RELU({{W}_{\text{local},j}}{{X}_{\text{masked}}})\\
&\phantom{\,\,\,\,\,\,\,\,}= RELU({{W}_{\text{local},j,1}}{{X}_{\text{masked},1}}+{{W}_{\text{local},j,2}}{{X}_{\text{masked},2}})\\
&{X}_{\text{global},j} = RELU({{W}_{\text{global},j}}{{X}_{\text{local},j}})\\
&X_{\text{globavg},j} = \frac{1}{n}\sum\limits_{k}{{{X}_{\text{global},j,k}}}\\&\phantom{\,\,\,\,\,\,\,\,\,\,\,\,\,\,\,\,\,\,\,\,\,\,\,\,\,\,\,\,\,\,\,\,\,\,\,\,\,\,\,\,\,\,\,\,\,\,\,\,\,\,\,\,\,\,\,\,\,\,\,\,\,\,\,\,\,\,\,\,\,\,\,\,\,\,\,\,\,\,\,\,\,\,\,\,\,\,\,\,\,\,\,\,\,\,\,\,\,\,\,\,\,\,\,\,\,\,\,\,} j\in (1,2)\\
&\phantom{\,\,}y=\text{softmax}({{X}_{\text{globavg},1}},{{X}_{\text{globavg},2}})
\end{split}
\end{equation*}

Where $X_\text{features}$ is given by the output of the face descriptors. $X_\text{masked}$ are the dot product of the weights $W_\text{mask}$ and the features.
It should be noted that, as explained above, there are two feature maps $X_local$ that are the convolution of the two original images feature maps. Those two new feature maps will eventually be computed for the probabilities of the two different classes for kin and non-kin.\\
The weights $W_\text{mask}$=$w_{\scriptscriptstyle{1}} \ldots w_{\scriptscriptstyle{4096}}$ are shared between both images and use as feature enhancement or suppression. the $W_\text{local}$ are weights that are shared between two image-specific features in order to create a convolution map from the two image's feature-maps.

\subsection{Loss Function}
The loss function of the network composed from the regularization part of the weights and the loss of the training examples computed by the softmax part.
Following \cite{FeatureSel} we are separating the regularization term for the mask layer and the regularization term for all the other weights in the network.
The total loss of the network is given by:
\begin{equation*}
\rm{Loss}_{{\rm{total}}} = \rm{Loss}{_{{\rm{reg,classifier}}}} + \rm{Loss}_{{{\rm{reg,mask}}}} + \rm{Loss}_{{{\rm{softmax}}}}
\end{equation*}

The regularization term for the weights in the local and the global part of the classifier is given by the $l^2$ norm, and can be expressed as:
\begin{equation*}
\begin{split}
\rm{Loss}_{\rm{reg,classifier}}& =\\
&-\lambda\left[{\sum\limits_{i=1}^{4096}{{w^2}_{local,i}}+\sum\limits_{k=1}^{4096\cdot128}{{w^2}_{global,k}}} \right]
\end{split}
\end{equation*}

Following \cite{FeatureSel}, the regularization term for the mask weights is computed by the $l^1$ norm and scaled by a factor proportional to the size of the layer:
\begin{equation*}
\rm{Loss}{_{\rm{reg,mask}}} =  - \frac{\lambda }{d}\sum\limits_{t = 1}^{4096} {\left| {{w_t}} \right|}
\end{equation*}

Where $\lambda$ is the regularize value, and $d$ is the network size.
The loss function of the softmax, $\rm{Loss}_{\rm{softmax}}$, is given by the cross-entropy between the targets and computed outputs from the networks.

\subsection{Derivatives Computation}
For training the network we compute the derivatives of the different layers, the weights that need to be updated are:\\ ${{W}_{mask}},{{W}_{local,1}},{{W}_{local,2}},{{W}_{global,1}},{{W}_{global,2}}$. To simplify, we will omit the derivatives for the VGG network and the derivatives for the regularization terms.\\
Following the Architecture described above and using the notation as in figure \ref{fig:total_arc}, the backward-pass gradients of the network can be computed by:

\begin{equation*}
\frac{\partial{{L}_{\text{soft}}}}{\partial{{X}_{\text{globavg}{_{1}}}}} = {{{\hat{y}}}_{1}}-{{y}_{1}}
\end{equation*}

\begin{equation*}
\frac{\partial {{L}_{\text{soft}}}}{\partial {{X}_{\text{global}{_{1}}}}}=\frac{\partial {{L}_{\text{soft}}}}{\partial {{X}_{\text{globavg}{_{1}}}}}\frac{\partial {{X}_{\text{globavg}{_{1}}}}}{\partial {{X}_{\text{global}{_{1}}}}}=({{{\hat{y}}}_{1}}-{{y}_{1}})\left[ \frac{1}{n}..\frac{1}{n} \right]
\end{equation*}

\begin{equation*}
\begin{split}
&\frac{\partial {{L}_{\text{soft}}}}{\partial {{W}_{\text{global},1,l,m}}} = \frac{\partial {{L}_{\text{soft}}}}{\partial {{X}_{\text{global}{_{1}}}}}\frac{\partial {{X}_{\text{global}{_{1}},(1\ldots t\ldots 128)}}}{\partial {{W}_{\text{global},1,l,m}}} \\ &\phantom{\,\,\,\,\,\,\,\,\,\,}=\frac{\partial {{L}_{\text{soft}}}}{\partial {{X}_{\text{global}{_{1}}}}}\left\{ \begin{matrix}
   {{X}_{\text{local},1,{{m}_{if\,\left( {{W}_{\text{global},1,t}}{{X}_{\text{local},1}}>0 \right)}}}}\,\,\,l=t  \\
   0\,\,\,\,\,\,\,\,\,\,\,\,\,\,\,\,\,\,\,\,\,\,\,\,\,\,\,\,\,\,\,\,\,\,\,\,\,\,\,\,\,\,\,\,\,\,\,\,\,\,\,\,\,\,\,\,\,\,\,\,\,\,\,\,\,\,\,\,\,l\ne t
\end{matrix} \right.
\end{split}
\end{equation*}

\begin{equation*}
\begin{split}
\frac{\partial {{L}_{\text{soft}}}}{\partial {{X}_{\text{local},1,l}}} = &\frac{\partial {{L}_{\text{soft}}}}{\partial {{X}_{\text{global}{_{1}}}}}\frac{\partial {{X}_{\text{global}{_{1}}(1\ldots t\ldots 128)}}}{\partial {{X}_{\text{local},1,l}}}\\ = &\frac{\partial {{L}_{\text{soft}}}}{\partial {{X}_{\text{global}{_{1}}}}}{{W}_{\text{global},1,t,{{l}_{if\,\left( {{W}_{\text{global},1,t}}{{X}_{\text{local},1}}>0 \right)}}}}
\end{split}
\end{equation*}

\begin{equation*}
\frac{\partial {{L}_{\text{soft}}}}{\partial {{W}_{\text{local},1}}}=\left[ \frac{\partial {{L}_{\text{soft}}}}{\partial {{W}_{\text{local},1,i}}},\frac{\partial {{L}_{\text{soft}}}}{\partial {{W}_{\text{local},1,j}}} \right] 
\end{equation*}

\begin{equation*}
\begin{split}
\frac{\partial {{L}_{\text{soft}}}}{\partial {{W}_{\text{local},1,i,l}}} = &\frac{\partial {{L}_{\text{soft}}}}{\partial {{X}_{\text{local},1}}}\frac{\partial {{X}_{\text{local},1,(1\ldots t\ldots 4096)}}}{\partial {{W}_{\text{local},1,i,l}}}\\ &\left\{ \begin{matrix}
   {{X}_{\text{masked},i,{{l}_{if\,({{W}_{\text{local},1,t}}{{X}_{\text{masked}}}>0)}}}}\,\,\,\,\,l=t  \\
   0\,\,\,\,\,\,\,\,\,\,\,\,\,\,\,\,\,\,\,\,\,\,\,\,\,\,\,\,\,\,\,\,\,\,\,\,\,\,\,\,\,\,\,\,\,\,\,\,\,\,\,\,\,\,\,\,\,\,\,\,\,\,\,\,\,\,\,\,\,l\ne t  \\
\end{matrix} \right.\\
\frac{\partial {{L}_{\text{soft}}}}{\partial {{W}_{\text{local},1,j,l}}} = & \frac{\partial {{L}_{\text{soft}}}}{\partial {{X}_{\text{local},1}}}\frac{\partial {{X}_{\text{local},1,(1\ldots t\ldots 4096)}}}{\partial {{W}_{\text{local},1,j,l}}}\\ &\left\{ \begin{matrix}
   {{X}_{\text{masked},j,{{l}_{if\,({{W}_{\text{local},1,t}}{{X}_{\text{masked}}}>0)}}}}\,\,\,\,\,l=t  \\
   0\,\,\,\,\,\,\,\,\,\,\,\,\,\,\,\,\,\,\,\,\,\,\,\,\,\,\,\,\,\,\,\,\,\,\,\,\,\,\,\,\,\,\,\,\,\,\,\,\,\,\,\,\,\,\,\,\,\,\,\,\,\,\,\,\,\,\,\,\,l\ne t  \\
\end{matrix} \right.
\end{split}
\end{equation*}

\begin{equation*}
\begin{split}
&\frac{\partial {{L}_{\text{soft}}}}{\partial {{X}_{\text{masked},l}}} = \left[ \begin{matrix}
   \frac{\partial {{L}_{\text{soft}}}}{\partial {{X}_{\text{masked},i,l}}}  \\
   \frac{\partial {{L}_{\text{soft}}}}{\partial {{X}_{\text{masked},j,l}}}  \\
\end{matrix} \right] \\
    = &\left[ \begin{matrix}
   \frac{\partial {{L}_{\text{soft}}}}{\partial {{X}_{\text{local},1}}}\frac{\partial {{X}_{\text{local},1,(1..t..4096)}}}{\partial {{X}_{\text{masked},i,l}}}+\frac{\partial {{L}_{\text{soft}}}}{\partial {{X}_{\text{local},2}}}\frac{\partial {{X}_{\text{local},2,(1..t..4096)}}}{\partial {{X}_{\text{masked},i,l}}}  \\
   \frac{\partial {{L}_{\text{soft}}}}{\partial {{X}_{\text{local},1}}}\frac{\partial {{X}_{\text{local},1,(1..t..4096)}}}{\partial {{X}_{\text{masked},j,l}}}+\frac{\partial {{L}_{\text{soft}}}}{\partial {{X}_{\text{local},2}}}\frac{\partial {{X}_{\text{local},2,(1..t..4096)}}}{\partial {{X}_{\text{masked},j,l}}}  \\
\end{matrix} \right] \\ 
 = &\left[ \begin{matrix}
   \left\{ \begin{matrix}
   \frac{\partial {{L}_{\text{soft}}}}{\partial {{X}_{\text{local},1}}}{{W}_{\text{local},1,i,{{t}_{(')}}}}+\frac{\partial {{L}_{\text{soft}}}}{\partial {{X}_{\text{local},2}}}{{W}_{\text{local},2,i,{{t}_{('')}}}}l=t  \\
   0\,\,\,\,\,\,\,\,\,\,\,\,\,\,\,\,\,\,\,\,\,\,\,\,\,\,\,\,\,\,\,\,\,\,\,\,\,\,\,\,\,\,\,\,\,\,\,\,\,\,\,\,\,\,\,\,\,\,\,\,\,\,\,\,\,\,\,\,\,\,\,\,\,\,\,\,\,\,\,\,\,\,\,\,\,\,\,\,\,\,\,\,\,\,\,\,\,\,\,\,l\ne t  \\
\end{matrix} \right.  \\
   \left\{ \begin{matrix}
   \frac{\partial {{L}_{\text{soft}}}}{\partial {{X}_{\text{local},1}}}{{W}_{\text{local},1,j,{{t}_{(')}}}}+\frac{\partial {{L}_{\text{soft}}}}{\partial {{X}_{\text{local},2}}}{{W}_{\text{local},2,j,{{t}_{('')}}}}l=t  \\
   0\,\,\,\,\,\,\,\,\,\,\,\,\,\,\,\,\,\,\,\,\,\,\,\,\,\,\,\,\,\,\,\,\,\,\,\,\,\,\,\,\,\,\,\,\,\,\,\,\,\,\,\,\,\,\,\,\,\,\,\,\,\,\,\,\,\,\,\,\,\,\,\,\,\,\,\,\,\,\,\,\,\,\,\,\,\,\,\,\,\,\,\,\,\,\,\,\,\,\,\,l\ne t\\
\end{matrix} \right.  \\
\end{matrix} \right] \\
&' =\rm{IF}\,({{W}_{\text{local},1,t}}{{X}_{\text{masked}}}>0) \\
&'' =\rm{IF}\,({{W}_{\text{local},2,t}}{{X}_{\text{masked}}}>0)
\end{split}
\end{equation*}

\begin{equation*}
\begin{split}
& \frac{\partial{{L}_{\text{soft}}}}{\partial{{W}_{\text{mask},l}}} =\frac{\partial{{L}_{\text{soft}}}}{\partial{{X}_{\text{masked},i}}}\frac{\partial{{X}_{\text{masked},i,(1\ldots t\ldots 4096)}}}{\partial{{W}_{\text{mask},l}}}\\
& \phantom{=\sum_{l=0}^{\infty}\,\,\,\,\,\,\,\,\,\,\,\Biggl(}  +\frac{\partial{{L}_{\text{soft}}}}{\partial{{X}_{\text{masked},j}}}\frac{\partial{{X}_{\text{masked},j,(1\ldots t\ldots 4096)}}}{\partial{{W}_{\text{mask},l}}}\\
& = \Biggl\{ \begin{matrix}
\frac{\partial {{L}_{\text{soft}}}}{\partial {{X}_{\text{masked},i}}}{{X}_{\text{feature},i,t}}+\frac{\partial {{L}_{\text{soft}}}}{\partial {{X}_{\text{masked},j}}}{{X}_{\text{feature},j,t}}\,\,\,\,\,l=t\\
0\,\,\,\,\,\,\,\,\,\,\,\,\,\,\,\,\,\,\,\,\,\,\,\,\,\,\,\,\,\,\,\,\,\,\,\,\,\,\,\,\,\,\,\,\,\,\,\,\,\,\,\,\,\,\,\,\,\,\,\,\,\,\,\,\,\,\,\,\,\,\,\,\,\,\,\,\,\,\,\,\,\,\,\,\,\,\,\,\,\,\,\,\,\,\,\,\,\,l\ne t
\end{matrix}
\end{split}
\end{equation*}

\section{Training Details}\label{ImpDet}
In this section, we will describe the different hyper-parameters that were chosen for training the network. we will continue with elaborate the different augmentation of the input data.
The pseudo-code for the training algorithm can be seen in algorithm \ref{alg:euclid}

\begin{algorithm}[H]
\caption{Training Algorithm}\label{alg:euclid}
\begin{flushleft}
\hspace*{\algorithmicindent} \textbf{Input: }\text{batch of image pairs}
\end{flushleft}
\begin{algorithmic}[1]
\Procedure{data augmentation}{}
\For{\text{im in images}}
\State $j\gets \text{rand(1,5)}$
\Switch{$j$}
   \Case{$1$}
      $\rm{im\_n}_i\gets \gamma(\rm{im},2)$
    \EndCase
    \Case{$2$}
      $\rm{im\_n}_i\gets \gamma(\rm{im},\frac{1}{2})$
    \EndCase
    \Case{$3$}
      $\rm{im\_n}_i\gets \rm{flip}(\rm{im})$
    \EndCase
    \Case{$4$}
      $\rm{im\_n}_i\gets \rm{flip}(\rm{im}),\gamma(\rm{im},2)$
    \EndCase
    \Case{$5$}
      $\rm{im\_n}_i\gets \rm{flip}(\rm{im}),\gamma(\rm{im},\frac{1}{2})$
    \EndCase
  \EndSwitch
  \EndFor
\State \textbf{return} $\rm{im\_n}$
\EndProcedure
\Procedure{face descriptors}{}
\State $X_{\rm{features},(1,2)}\gets VGG_{\rm{FACE}}(\rm{image}_1,\rm{image}_2)$
\State \textbf{return} $X_{\rm{features},(1,2)}$
\EndProcedure

\Procedure{forward pass}{}
\State $X_{\rm{masked},(1,2)}\gets W_{\rm{mask}}\cdot X_{\rm{features},(1,2)}$
\State $X_{\rm{local},(1,2)}\gets\rm{conv} (X_{\rm{masked},(1,2)},W_{\rm{local}})$
\State $X_{\rm{global},(1,2)}\gets\rm{fc} (X_{\rm{local},(1,2)},W_{\rm{global}})$
\State $X_{\rm{average},(1,2)}\gets\frac{1}{n}\sum\limits_{k}{{{X}_{\rm{global},(1,2),k}}}$
\State $y\gets\rm{softmax}({{X}_{\rm{average},1}},{{X}_{\rm{average},2}})$
\State $\rm{L}_{\rm{soft}}\gets-{(y'\log(y) + (1 - y')\log(1 - y))}$
\State $\rm{L}_{\rm{class}}\gets-\lambda\left[l^2_{\rm{norm}}(W_{\rm{local}})+l^2_{\rm{norm}}(W_{\rm{global}})\right]$
\State $\rm{L}_{\rm{mask}}\gets-\frac{\lambda}{d}l^1_{\rm{norm}}(W_{\rm{mask}})$
\State $\rm{Loss}_{\rm{total}}\gets L_{\rm{soft}} + L_{\rm{class}} + L_{\rm{mask}}$
\State \textbf{return} $\rm{Loss}_{\rm{total}}$
\EndProcedure

\Procedure{backward pass}{}
\State \textbf{compute} $\frac{\partial {{L}_{\rm{total}}}}{\partial {{W}_{\rm{global}}}},\frac{\partial {{L}_{\rm{total}}}}{\partial {{W}_{\rm{local}}}},\frac{\partial {{L}_{\rm{total}}}}{\partial {{W}_{\rm{global}}}}$
\State ${W}_{mask,new}\gets{W}_{mask} + lr\frac{\partial {{L}_{\rm{total}}}}{\partial {{W}_{\rm{mask}}}}$
\State ${W}_{local,new}\gets{W}_{local} + lr\frac{\partial {{L}_{\rm{total}}}}{\partial {{W}_{\text{local}}}}$
\State ${W}_{global,new}\gets{W}_{global} + lr\frac{\partial {{L}_{\rm{total}}}}{\partial {{W}_{\rm{global}}}}$
\State \textbf{return} ${W}_{mask,new},{W}_{local,new},{W}_{global,new}$
\EndProcedure
\end{algorithmic}
\end{algorithm}

\subsection{Network Parameters}
\begin{figure}[h]
\includegraphics[height=2.6in, width=3in]{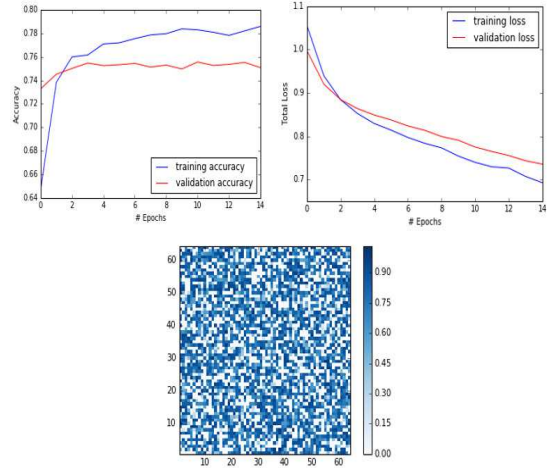}
\caption{Example of the training results for sisters verification, upper left is the training accuracy, upper right is the training loss and in the bottom is the resulting masking layer}
\label{fig:loss_acc}
\end{figure}

The face descriptor is implemented by the Keras version of VGG-FACE with depth $16$ as suggested in \cite{keras_vgg}.
VGG has an architecture of $16$ layers, composed of convolution layers, max pooling, and finally fully connected layers.
The output of this network is a classification for $1000$ classes. We used the output of the convolution layers (i.e., the $4096$ embeddings).
The VGG-FACE was pre-trained with one million celebrities dataset.\\ 
As described in \ref{Dev}, we proposed two regularization terms, one for the self-weighting layer, and the second for the local-global classifier.
For the first regularization term we used $l^1$ regularization with $\lambda=0.5$ and $d=4,096$.
For the second regularization term we used $l^2$ regularization with $\lambda=1e-5$\\
We Trained our network using Adam optimizer with a learning rate of $1e-4$ to $1e-5$ and $beta_1=0.9, beta_2=0.999, epsilon=1e-08, decay=0.0$.
We used dropout layer with $p=0.8$ between the local part and the global part of the classifier.\\
For choosing those hyper-parameters we trained the different classifiers using the training set and validate it with the validation set, we stopped the training when the validation accuracy stopped raising. An example for the training results and the resulting mask layer can be seen in figure  \ref{fig:loss_acc}. For the Final submission, we trained our model with the train and validation set.\\
We initialize the weights for the different layers of the network with random numbers. And initialize all the weights of the mask layer as $1$ since we want the network to self-adjust the weights for each of the features.

\subsection{Data Augmentation}
\begin{figure}[h]
\includegraphics[height=2.6in, width=3.2in]{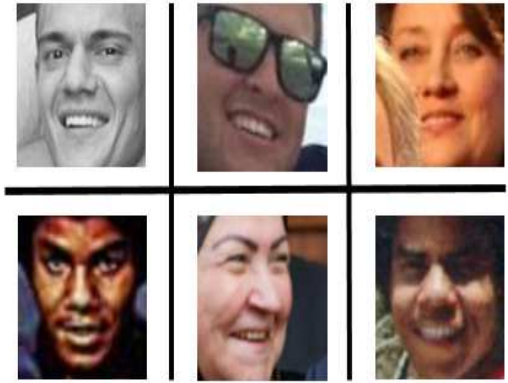}
\caption{Example for the diversity of the RFIW dataset in color, pose and illumination}
\label{fig:diversity_dat}
\end{figure}

Although the Data that was released for the RFIW2018 competition is the most massive kinship data available, still, it is proven that augmenting the face data can improve the performance of face recognition task. Further, when investigating the RFIW2018 data one can see the diversity in the face images, such as, pose, illumination, color, etc. This diversity can help to train a better and general network.
An example of this diversity can be seen in figure \ref{fig:diversity_dat}.\\
For each image we randomly select the augmentation method it will pass:
\renewcommand{\labelenumii}{\roman{enumi}}
 \begin{enumerate}
   \item randomly change the gamma factor of the image for different illumination, scale between $\gamma\in(0.5,1,2)$.   
   \item horizontal flipping the image.
 \end{enumerate}
An augmentation for a single image can be any combination of one or more of the listed above.

\section{Experimental Result}
\label{Results}
In this section, we describe our result based on the challenge of RFIW2018.
We start by describing the dataset that was used. We then continue with describing our results on the different cases that ware trained. We conclude our work and suggesting new areas for research.

\begin{table}[h]
  \caption{FIW Dataset Description and Details}
  \label{tab:dataset}
  \begin{tabular}{lccc}
    \toprule
    Type/Phase & Train & Validation & Test \\
    \midrule
    Brothers & 29,812 & 55,546 & 18,196 \\
    Sisters & 19,778 & 35,024 & 4,796 \\
    Siblings & 28,428 & 15,422 & 9,716 \\
    Father-Daughter & 41,604 & 35,238 & 15,040 \\
    Father-Son & 64,826 & 44,870 & 18,166 \\
    Mother-Daughter & 39,110 & 29,012 & 14,394 \\
    Mother-Son & 66,464 & 31,094 & 14,806 \\
    Grfather-Grdaughter & 1,478 & 4,846 & 838 \\
    Grfather-Grson & 1,388 & 1,926 & 1,588 \\
    Grmother-Grdaughter & 1,580 & 3,768 & 952 \\
    Grmother-Grson & 1,284 & 1,844 & 1,470 \\
    
  \bottomrule
  Total & 295,752 & 258,590 & 99,962 \\
\end{tabular}
\end{table}

\begin{table*}[t]
\centering
  \caption{Results for different models trained with self-adjust weights layer and without}
  \label{tab:Results}
  \begin{adjustbox}{width=1\textwidth}

      \begin{tabular}{l cccccccccccc}
        \toprule
        Model/Relation & BB & SS & SIBS & FD & FS & MD & MS & GFGD & GFGS & GMGD & GMGS & Average \\
        \midrule
        With Self-adjusted & 0.6914 & 0.7725 & 0.6864 & 0.6891 & 0.6889 & 0.7378 & 0.7085 & 0.6193 & 0.6366 & 0.6386 & 0.6319 & 0.6819 \\
        With Self-adjusted t.h. at half  & 0.6946 & 0.7733 & 0.6876 &  0.6890 & 0.6846 & 0.7389 & 0.7026 & 0.6229 & 0.6234 & 0.6344 & 0.6285 & 0.6800 \\
        W/O Seld-adjusted & 0.7060 & 0.7723 & 0.6865 & 0.6232 & 0.6961 & 0.7348 & 0.7172 & 0.5763 & 0.6114 & 0.6523 & 0.6346 & 0.6737 \\\hline
    \end{tabular}
\end{adjustbox}
\end{table*}

\begin{table*}[h]
\centering
  \caption{Results from the challenge site}
  \label{tab:SiteRes}
  \begin{adjustbox}{width=1\textwidth}
    \begin{tabular}{l cccccccccccc}
        \toprule
        Name/ Relation & MD & MS & SS & BB & SIBS & GMGD & GMGS & FS & GFGS & FD & GFGD & Average\\
        \midrule
        eranda-ours & 0.73794 & 0.70856 & 0.77251 & 0.69147 & 0.68649 & 0.63865 & 0.63265 & 0.68897 & 0.63665 & 0.68922 & 0.61933 & 0.68204 \\
        Unknown & 0.66625 & 0.59840 & 0.72852 & 0.66256 & 0.62968 & 0.57563 & 0.61088 & 0.61527 & 0.56927 & 0.63304 & 0.57995 & 0.62449 \\
        Unknown & 0.65888 & 0.60468 & 0.70204 & 0.68443 & 0.61620 & 0.56932 & 0.62789 & 0.61224 & 0.56360 & 0.63071 & 0.59427 & 0.62402 \\
        Unknown & 0.59552 & 0.57814 & 0.64678 & 0.57221 & 0.57400 & 0.55357 & 0.57687 & 0.57420 & 0.55667 & 0.59355 & 0.55966 & 0.58011 \\
        Unknown & 0.61553 & 0.60482 & 0.68015 & 0.62700 & 0.59016 & 0.54516 & 0.52176 & 0.58152 & 0.53274 & 0.58211 & 0.57875 & 0.58725 \\
        Unknown & 0.56704 & 0.54086 & 0.60196 & 0.51319 & 0.55393 & 0.52416 & 0.50204 & 0.54123 & 0.52455 & 0.55957 & 0.55131 & 0.54362 \\\hline
    \end{tabular}
\end{adjustbox}
\end{table*}

\subsection{Recognize Families in the Wild 2018}
The first track of Recognize Families In the Wild 2018 (RFIW2018) challenge deals with solving a classifier for kinship verification to $11$ types of relationships as shown in \ref{tab:dataset}.
The FIW (families in the wild) dataset contains images of $1,000$ different family-cells, for a total of $11,932$ images. The images are divided into $11$ family relations as mentioned above and arranged as positive and negative examples.
Each one of the family members has a variant amount of images from different ages. The total amount of pairs (positive and negative) is $654,304$. \\
The images also contain meta-data for gender. The images are characterized with a different background, face-color, size of the face in the images, pose and expressions.
The FIW dataset is the only dataset that also contains face images and relations for grandfather, grandmother/grandson, granddaughter.\\
The competition lasted for a defined period and was divided into different phases, in the first phase, training phase, the training data is released, then in the validation phase, the model trained is validated using a new validation data. Finally, in the test phase, the model is tested with unseen test data.

\subsection{Test Results and Discussion}
During training we optimized our architecture in the sense of choosing the different hyper-parameters for regularization terms as described in \ref{ImpDet}, choosing the size for the fully connected layers and selecting the dropout probability.
In this section, we will show the results of the optimization. We can distinguish the results from three experiments, training, and testing with the self-adjust layer, without it, and with threshold the self-adjusted layer by half.\\
\textbf{Without self-adjusted layer} - In this case, we trained our network with only updating the weights for the local-global classifier, and tested the test data with the networked trained.\\
\textbf{With self-adjusted layer} - In this case we trained our model, end to end, with the self-adjusted layer, allowing it to change the impact of each feature according to its contribution to the kinship-verification task. The weights of the local-global classifier also were modified during training.
When testing the network, we used the trained weights for the self-adjusted layer.\\
\textbf{With self-adjusted layer threshold at half} - In this case, we did the same training procedure as in with self-adjusted layer, while in the testing phase, we chose a threshold value that divides the features to half(median). we used this threshold to cancel all the features that were below the threshold and with this, also canceling all the weights that were connected to those original features.
The results of the above description are shown in table \ref{tab:Results}.\\
We show that there is no significant change (about $1$ percent on average) between the two models trained.
We saw that the potential of this self-adjusted layer is that by training this layer, we have the possibility to reduce the number of features by a factor of $2$ while not decreasing the performance of the network, and on average, the network is still better than a network that was not trained with this selecting layer.
It is noticeable that the average performance is reduced because of the poor performance of the GFGD, GFGS, GMGD, GMGS pairs which come with coherence to the lack of data for those pairs.
We conclude with presenting table \ref{tab:SiteRes} - the leader bored from RFIW challenge.

\section{CONCLUSIONS AND FUTURE WORKS}
\label{CONCLUSION}

In this work, we aimed to solve the Kinship Verification problem with the novel method of a deep learning network.
We presented an architecture that purposes to compete for the strict usage of kinship verification, without having the ID's of the people in the dataset. Moreover, not having the specific family classification for each photo, so we do not have the knowledge during training to select negative or positive examples that were not being given in the training set.\\
We presented a layer that self-adjust weights of face features for the task of kinship verification.
We show that we can use the self-adjusted weights to reduce the number of weights used in the network without the loss of performance.\\
We introduced the use of local and global classification, explaining why it is needed in combining two feature maps created for face recognition, we developed the derivatives for the network and this classifier as well.
Finally, we showed that our model wins the RFIW2018 (FG2018) challenge.\\
For future work, one can try using a more general face feature extractor with different architecture then VGG-16. one can also try to train the network with the relax scenario presented in the competition (use the ID's of people in the dataset for refining the face feature extractor or for creating more sophisticated examples from the datast)
it is also optional to use the self-adjusted weights layer, with a different purpose such as making the model invariant to age or color diversity in the dataset. 
\FloatBarrier

\bibliographystyle{unsrt}
\bibliographystyle{ieeetr}
\nocite{*}\bibliography{sample-bibliography} 

\end{document}